\documentclass[10pt, a4paper]{article}
\usepackage{lrec-coling2024} % this is the new style
\usepackage{inconsolata}
\usepackage{tabularx}
\usepackage{amsthm,amsmath,amssymb}
\usepackage{mathrsfs}
\usepackage{graphicx}
\usepackage{multicol,multirow}
\usepackage{algorithm, algorithmic}
\usepackage{booktabs}
\usepackage{subcaption}
\usepackage{makecell}

\title{Continual Few-shot Event Detection via Hierarchical Augmentation Networks}

\name{Chenlong Zhang$^{1,2}$\sthanks{\quad These authors contribute equally to this work.}, Pengfei Cao$^{1,2}$\footnotemark[1], Yubo Chen$^{1,2}$\sthanks{\quad Corresponding author.}, Kang Liu$^{1,2,3}$
\\ {\bf \large Zhiqiang Zhang$^{4}$, Mengshu Sun$^{4}$, Jun Zhao$^{1,2}$}
}

\address{
        $^{1}$The Laboratory of Cognition and Decision Intelligence for Complex Systems,\\
        Institute of Automation, Chinese Academy of Sciences, Beijing, China\\ 
        $^{2}$School of Artificial Intelligence, University of Chinese Academy of Sciences, Beijing, China\\
        $^{3}$Shanghai Artificial Intelligence Laboratory, Shanghai, China\\
        $^{4}$Ant Group, Hangzhou, China\\
         zhangchenlong2023@ia.ac.cn\\
         \{pengfei$.$cao, yubo$.$chen, kliu, jzhao\}@nlpr.ia.ac.cn \\
        }

\abstract{
Traditional continual event detection relies on abundant labeled data for training, which is often impractical to obtain in real-world applications. In this paper, we introduce continual few-shot event detection (CFED), a more commonly encountered scenario when a substantial number of labeled samples are not accessible. The CFED task is challenging as it involves memorizing previous event types and learning new event types with few-shot samples. To mitigate these challenges, we propose a memory-based framework: \textbf{H}ierarchical \textbf{A}ugmentation \textbf{Net}works (\textbf{HANet}). To memorize previous event types with limited memory, we incorporate \textit{prototypical augmentation} into the memory set. For the issue of learning new event types in few-shot scenarios, we propose a \textit{contrastive augmentation} module for token representations. Despite comparing with previous state-of-the-art methods, we also conduct comparisons with ChatGPT. Experiment results demonstrate that our method significantly outperforms all of these methods in multiple continual few-shot event detection tasks. 
 \\ \newline  
 }

\begin{document}

\maketitleabstract

\section{Introduction}
\begin{figure}[t]
    \centering
    \includegraphics[width=\linewidth]{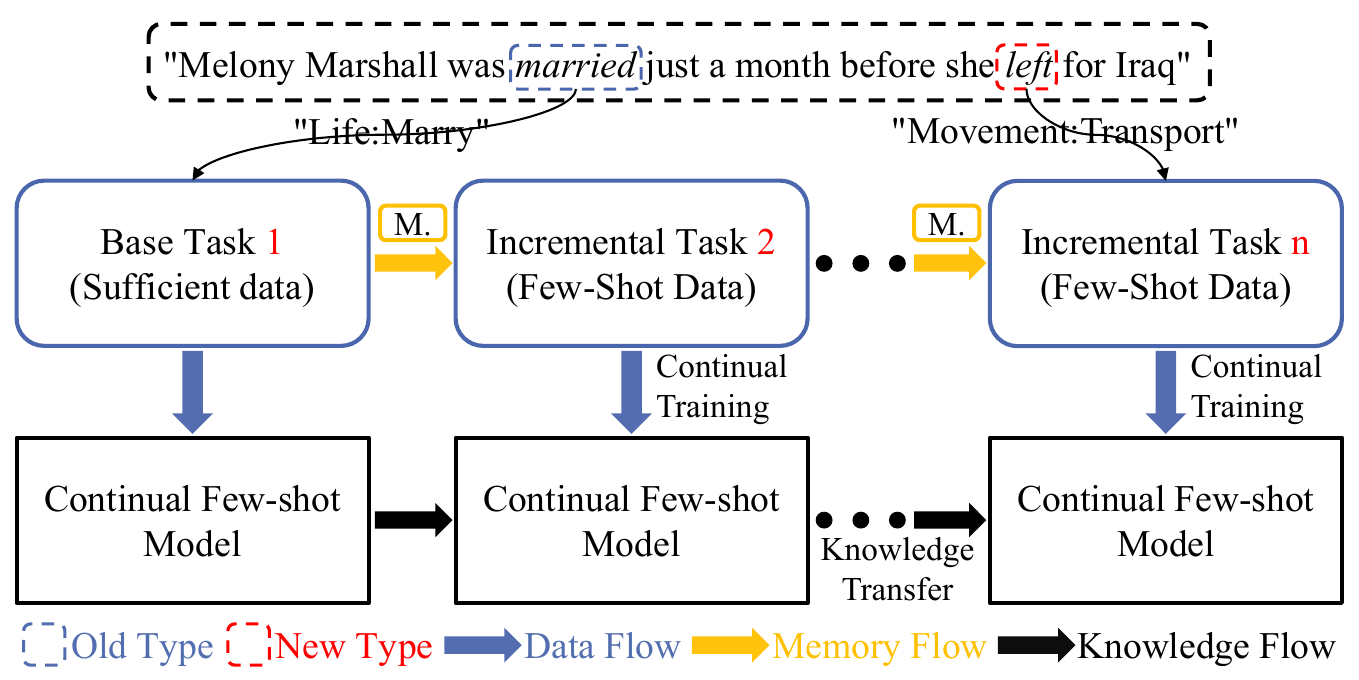}
    \caption{Memory-based framework for continual few-shot event detection. It preserves previous knowledge by maintaining a memory set ``M.'' and transferring knowledge from previous models.}
  
    \label{fig_task}
\end{figure}

\textbf{Event Detection} (ED) involves detecting event triggers and classifying the corresponding event types \citep{ahn2006stages} (e.g., in Figure \ref{fig_task}, the words ``married'' and ``left'' trigger events ``Marry'' and ``Transport'', respectively.). It is an essential information extraction task that can be applied in various natural language processing applications. Conventional methods \citep{chen-etal-2015-event,nguyen-grishman-2015-event} commonly model ED as a supervised task trained on fixed data with pre-defined event types. However, in real-world applications, new event types emerge continually.

Thus, \textbf{Continual Event Detection} (CED) has been proposed \citep{cao-etal-2020-incremental, yu-etal-2021-lifelong}. The CED task assumes multiple ED tasks emerge continually, which requires ED models to learn new types while maintaining the capability of detecting previous types. The CED task is challenging due to the catastrophic forgetting problem \citep{mccloskey1989catastrophic}, where the model's performance on previous tasks declines significantly when learning new tasks. To mitigate such a dilemma, previous works have proved that memory-based methods (see Figure \ref{fig_task}) are the most effective in solving CED task \citep{cao-etal-2020-incremental,yu-etal-2021-lifelong,liu-etal-2022-incremental}. These methods preserve prototypical samples as memory set to replay previous knowledge. Abundant representative features can effectively remind the model of previous types, achieving state-of-the-art performance.

Even though these methods achieve remarkable performance, they all assume that the training samples in incremental tasks are sufficient. Actually, in practical applications, new events emerge successively, making it infeasible to obtain a sufficient number of high-quality samples for each emerging new event type. It is more commonplace to encounter incremental tasks with only a handful of annotated samples (e.g., 10, 5, or even 1) for each new type. Nonetheless, this circumstance has been overlooked by previous works. 

To this end, we propose a new task: \textbf{Continual Few-shot Event Detection} (CFED), which aims to continually learn new event detection tasks with few-shot samples. For example, as shown in Figure \ref{fig_task}, the first task (base task) denotes the regular ED task with abundant training samples (e.g., 100 samples are available for event type ``Life: Marry''). Then, only a few samples are available for the emerging incremental tasks (e.g., there are only 5 labeled samples accessible for the new type ``Movement: Transport''). 

Obviously, CFED introduces a more challenging yet realistic scenario as it requires \textit{memorizing previous event types} and \textit{learning new event types} with few-shot samples. We present the two challenges specifically as follows:

\textbf{Memorizing previous event types with few-shot samples}: In the CED task, memroy-based methods use a multitude of exemplars (e.g., 50) in memory set to effectively characterize the prototypical feature space, thus alleviating catastrophic forgetting. However, in the CFED task, only 10, 5, or 1 sample is available for training. In extreme scenarios, there is only one sample per type available to be stored in the memory set for further replay. Therefore, how to utilize rare stored samples to mitigate catastrophic forgetting remains challenging.

\textbf{Learning new event types with few-shot samples}: 
Supervised methods usually require a large number of annotated samples \citep{lai-etal-2020-extensively,deng2020meta,zhang-etal-2022-hcl}. When trained with limited samples, these methods often struggle to generalize well and suffer from overfitting. Current large language models (llms) \citep{brown2020language,touvron2023llama} have demonstrated promising capability to learn from few-shot samples with their in-context learning ability. However, these models are constrained by limited knowledge (e.g., ChatGPT's knowledge of world and events is limited after 2021). Though in-context learning is capable of temporarily empowering them with new event knowledge, it fails to truly inject this knowledge into the model\citep{moiseev-etal-2022-skill}. Therefore, We consider using a fine-tuned language model to solve the CFED task. How to effectively mitigate overfitting with few-shot samples for learning new event types is still a formidable challenge.

To address these problems, we propose a memory-based approach: \textbf{H}ierarchical \textbf{A}ugmentation \textbf{Net}work (\textbf{HANet}). When memorizing previous types, we devise \textit{prototypical augmentation} to augment the prototypical feature space of exemplars in the memory, thus alleviating catastrophic forgetting. To address overfitting in learning new types, we design \textit{contrastive augmentation} module to acquire valuable information from few-shot samples. Experimental results show that our method surpasses previous baselines significantly. 

Our contributions can be summarized as follows: 

(1) To the best of our knowledge, we are the first to propose continual few-shot event detection and construct benchmarks based on ACE and MAVEN. 

(2) We propose a \textbf{H}ierarchical \textbf{A}ugmentation \textbf{Net}work (\textbf{HANet}), which leverage prototypical augmentation and contrastive augmentation to memorize previous event types and to learn new event types with few-shot samples. 

(3) Experimental results demonstrate that our method significantly outperforms previous state-of-the-art methods in all CFED settings. Impressively, our method achieves $7.27\%$ and $8.44\%$ improvements on micro F1 in 4-way 5-shot MAVEN and 2-way 5-shot ACE settings. Moreover, experiments with ChatGPT show that our method achieves superior results. Our code and dataset are publicly available at \url{https://github.com/chenlong-clock/CFED-HANet}.
 \begin{figure*}[t]
    \centering
    \includegraphics[width=\textwidth]{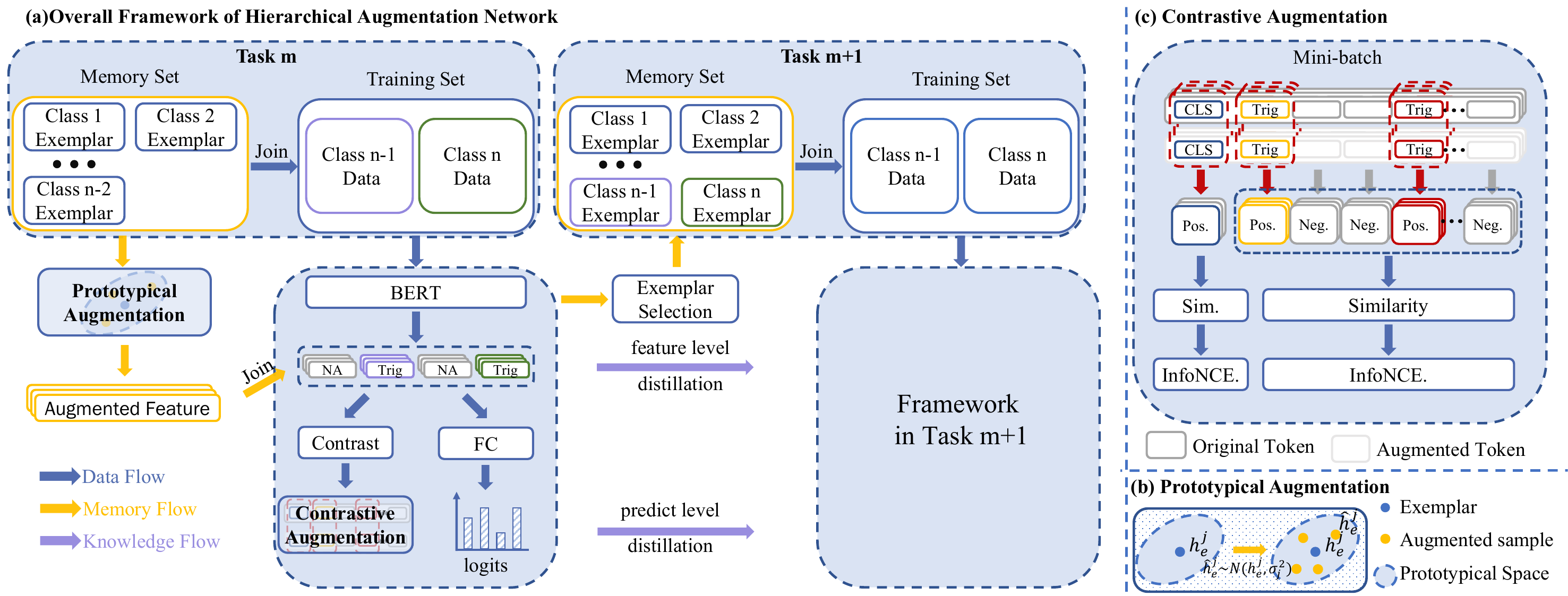}
    \caption{Our system consists of a general event detector, prototypical augmentation, and contrastive augmentation. When learning new tasks with an event detector, the model replays prior knowledge from the augmented feature. Then, contrastive augmentation maximizes the acquisition of knowledge from few-shot samples.}
    % 简要的介绍
    \label{fig_frame}
  
\end{figure*}

\section{Problem Definition} \label{section:definition}
Continual few-shot event detection (CFED) aims to detect emerging events with few-shot samples. As shown in Figure \ref{fig_task}, given tasks $\mathbb{T} = \{T_1, T_2, ..., T_n\}$, each task has individual training/validation/testing set $T_i = \{D_i^{train}, D_i^{dev}, D_i^{test}\}$. $D_i = \left\{(\mathbf{X}^j_i, \mathbf{Y}^j_i)\right\}^m_{j=1}$, where $\mathbf{X}$ and $\mathbf{Y}$ are samples and their corresponding labels, and $m$ is the number of event types in each task. The first sub-task $T_1$ is the base task $T_{base}$ that contains abundant training samples. The rest sub-tasks are defined as few-shot incremental tasks $T_{inc} = \{T_2, T_3, ..., T_n\}$, with only a few samples (e.g., 5 or 10) for each new event type. For any two tasks $T_i$ and $T_j$, their types are non-overlapping: $T_i \cap T_j = \emptyset$. At time step $t$, for CFED task $C_t$, the training set is formulated as $C_t^{train} = D_t^{train}$ and the validation/testing set is $C_t^{test} = D_t^{test} \bigcup C_{t-1}^{test}$, indicating the CFED system is supposed to keep stable performance on all observed labels $L_t=\bigcup_{i=1}^t \{\mathbf{Y}_i^j\}_{j=1}^m$ with the currently available training samples in task $T_t$. 

\section{Methodology}
The framework of our method is illustrated in Figure \ref{fig_frame}. It comprises a general \textit{event detector}, a memory enhanced by \textit{prototypical augmentation}, and a \textit{contrastive augmentation} module. For input sentences, \textit{event detector} performs trigger extraction. Then, the exemplars are augmented by \textit{prototypical augmentation} to replay previous knowledge. Additionally, contrastive augmentation exploits information from each sample by applying an auxiliary contrastive loss. We provide a detailed introduction as follows.
\subsection{Event Detector} \label{sec:ed}
The event detector is composed of a trigger extractor and a classifier. Following previous works \citep{cao-etal-2020-incremental,liu-etal-2022-incremental}, we implement a pre-trained 12-layer BERT \citep{devlin-etal-2019-bert} model to encode sentences. Specifically, given a sentence $\mathbf{S} = \{\mathbf{x}_1, \mathbf{x}_2, ..., [\mathbf{e}_s, ...,\mathbf{e}_e], ..., \mathbf{x}_n \}$ containing event triggers $\mathbf{E} = [\mathbf{e}_s, ..., \mathbf{e}_e]$, the hidden representation is $\mathbf{H}\in\mathbb{R}^{n \times d}$. We get hidden states of a trigger $\mathbf{H_e}$ by concatenating their start and end representations. Then, $p(\mathbf{y}_i|\mathbf{h}_e)$ for event type $\mathbf{y}_i \in L_t$ at stage $t$ is obtained by the following equation:
\begin{equation} 
    p(\mathbf{y}_i|\mathbf{h}_e)= \frac{\exp{(\mathbf{W}_i^T \mathbf{h}_e + \mathbf{b}_i)}}{\sum_{j=1}^{|L_t|} \exp{(\mathbf{W}_j^T \mathbf{h}_e + \mathbf{b}_j)}} \label{eq:dis}
\end{equation}
where $\mathbf{W}_i \in \mathbb{R}^{d \times |L_t|}$ is a linear projection for classification. The possible types are $L_t$. Then, we train the model with Cross Entropy Loss:
\begin{equation} 
    \mathcal{L}_{ce} = - \sum_{(X,Y) \in T_t} \mathbf{y} \log \mathbf{p} 
\end{equation}
where $\mathbf{y}$ is the ground-truth label for trigger $\mathbf{h}_e$, $\mathbf{p}$ is the label distribution calculated by Equation (\ref{eq:dis}).

\subsection{Prototypical Augmentation}
We construct a memory set by selecting the most representative examples. Accordingly, we adopt a distance-based algorithm. Finally, prototypical augmentation is applied in the feature space. 
\subsubsection{Memory Construction} \label{section:memory}
After task $T_t$, we combine a memory set $M_t$ comprising exemplars of current types with previous memory $M_{t-1}$. Since only few samples are available for training in incremental tasks, the most extreme condition should be taken into account so that our method can be compatible with any real-world applications. Thus, we only select one exemplar $(\mathbf{x}_{e,t}^j, \mathbf{y}_{e,t}^j)$ for every category in $T_t$:
\begin{equation} 
    M_t = 
    \begin{cases}
        \left\{(\mathbf{x}_{e,t}^j, \mathbf{y}_{e,t}^j)\right\}_{j=1}^m,&\text{if } t=1 \\
        \left\{(\mathbf{x}_{e,t}^j, \mathbf{y}_{e,t}^j)\right\}_{j=1}^m\bigcup M_{t-1},&\text{if } t>1
    \end{cases}
    \label{equation:memory}
\end{equation}
The combined $M_t$ is then treated as a part of the training set in the next task $T_{t+1}=T_{t+1} \bigcup M_t$. To select the most representative samples, we first create a prototype for each event type by averaging the encoded representations. Then we choose the closest sample measured by distance (e.g., \emph{$L_2$ Distance} or \emph{Cosine Distance}) as the exemplar.

\subsubsection{Prototypical Augmentation}
Since conventional memory preserves plenty of representative samples, these samples characterize the feature space of their types. However, in our settings, the memory is limited to 1 for each type. The exemplar can only be represented as a point in the feature space (see Figure \ref{fig_frame} (b)). To tackle this, we reconstruct the feature space of the exemplar by prototypical augmentation.
 
We get the exemplar's representation $\mathbf{h}_e^j$ that belongs to class $j$. We assume the pseudo feature space follows Gaussian Distribution. In view that exemplars are normally considered the most representative sample, their representation is regarded as the mean. The variance of the distribution is calculated in the exemplar selection process, where we calculate the mean squared deviation of all samples that belong to the same category:
\begin{equation} 
    \mathbf{\sigma}^2_j=\frac{1}{|\mathbf{H}_e^j|}\sum_{\mathbf{h}_i^j \in \mathbf{H}^j}(\mathbf{h}_i^j - \mathbf{\mu}_j)^2
\end{equation}
where $H^j_e$ are BERT representations that belong to event type $\mathbf{Y}_t^j$. According to Equation (\ref{equation:memory}), the memory set $M_{t-1}$ is reformulated as $M_{t-1} = \bigcup_{k=1}^{i-1} \left\{ (\mathbf{x}_{e,k}^j, \mathbf{y}_{e,k}^j, {\mathbf{\sigma}^2}_t^j) \right\}_{j=1}^m$. We define the mean squared deviation of all exemplars as the variants of Gaussian distribution. When replaying exemplars, given the representation of exemplar $\mathbf{h}_e^j$, we have $\mathbf{\mu}_j = \mathbf{h}_e^j$. Then, we sample from the distribution to construct synthetic features multiple times:
\begin{equation} 
    \hat{\mathbf{H}}^{j}_e = \{\hat{\mathbf{h}}^{j}_{e,1}, \dots, \hat{\mathbf{h}}^{j}_{e,n}\} \sim \mathcal{N}(\mathbf{\mu}_j, \mathbf{\sigma}^2_j)
\end{equation}
These synthetic features can represent the feature space of their category (i.e., prototypical space). Then we replay the memory:
\begin{equation} 
    \mathcal{L}_{re}= -\sum^{\hat{\mathbf{H}}^{j}_e} \mathbf{y}_j \log \hat{\mathbf{p}}_j
\end{equation}
where $\hat{\mathbf{p}}_j$ is obtained from $\mathbf{H}^{j}_e$ by Equation (\ref{eq:dis}).

\subsection{Contrastive Augmentation}
Overfitting is likely to appear in $T_{inc}$ when learning few-shot new event types. As shown in Figure \ref{fig_frame}(c), we propose contrastive augmentation (CA) to uncover the implicit inter-information in the token scale. Following \citet{zhang-etal-2022-new}, we use multiple data augmentations (e.g., Dropout, Random Token Shuffle, and Random Token Replacement) to generate augmented tokens. These tokens are used to construct positive and negative pairs. Finally, we propose two contrastive losses to aggregate the information.

\begin{algorithm}[t]
    \caption{Training procedure}
    \begin{algorithmic}[1]
        \REQUIRE Base task ${T_1}$, incremental task $\{T_2, ..., T_n\} $ and model's parameter $\theta$
        \STATE initialize $\theta_1$ for base task $T_1$
        \STATE update parameter $\theta_1$ in task $T_1$ using loss function $\mathcal{L}_{ce}$ and $\mathcal{L}_{cls}$
        \STATE get memory set $M_{1}$ from $T_1$ and $\theta_1$
        \FOR{$i=2$ to $n$} 
        \STATE get a copy of the previous model's parameter $\theta_{i-1}$
        \STATE freeze parameter $\theta_{i-1}$
        \STATE get combined training set $T_i = T_i \cup M_{i-1}$
        \STATE update parameter $\theta_i$ in task $T_i$ using loss function $\mathcal{L}_{ce}$, $\mathcal{L}_{fd}$, $\mathcal{L}_{pd}$, $\mathcal{L}_{re}$, $\mathcal{L}_{cls}$ and $\mathcal{L}_{trig}$
        \STATE get memory set $M_i$ from $T_i$ and $\theta_i$
        \STATE update memory set $M_i$ = $M_i \cup M_{i-1}$
        \ENDFOR

    \end{algorithmic}
    \label{alg:train}
\end{algorithm}

\subsubsection{Contrastive Pairs Construction}
We first construct positive pairs and negative pairs from batched data. Specifically, given a mini-batch $\mathcal{B}=\{(\mathbf{x}_i, \mathbf{y}_i)\}_{i=1}^n$, the original sentences are $(\mathbf{x}_i^1, \mathbf{y}_i^1)$ and the augmented sentences are $\{(\mathbf{x}_i^k, \mathbf{y}_i^k)\}_{k=2}^{m+1}$, where $m$ is a hyperparameter, denoting the augmentation times. Thus, sentences that have the same origin can be described as $\mathcal{O}=\{(\mathbf{x}_i^k, \mathbf{y}_i^k)\}_{k=1}^{m+1}$. Based on these pairs, we perform contrastive learning in sentence representation and trigger representation.
 \subsubsection{Contrastive Sentence Representation Learning}
As in BERT, the special [CLS] token generally conveys the sentence representation. Similar to \citet{mou-etal-2022-disentangled}, we utilize contrastive sentence representation learning for $\mathbf{h}_{cls}$. Representations originating from the same sentence are regarded as positive pairs and those that originate from different sentences are regarded as positive pairs. We leverage InfoNCE loss \citep{oord2018representation}:
\begin{equation} 
    \begin{aligned}
        &\mathcal{L}_{cls} = \frac{1}{n-1}\sum_{i}^{|\mathcal{B}|}  -\frac{1}{m} \sum_{j \neq k}^{|\mathcal{O}|} \\ 
        & \log \frac{\exp(S({\mathbf{h}_{cls}}_i^j,{\mathbf{h}_{cls}}_i^k)/\tau)}{\sum_{p \neq i}^{|\mathcal{B}|} \sum_{q}^{|O|} \exp(S({\mathbf{h}_{cls}}_i^j {\mathbf{h}_{cls}}_p^q)/\tau)} \\
    \end{aligned}
\end{equation}
where $S(\cdot)$ is the similarity function, and $\tau$ is a temperature parameter to smooth the distribution and control the similarity range by scaling the output.

\subsubsection{Contrastive Trigger Representation Learning}
Considering trigger representations, we propose to construct positive pairs when triggers within $\mathcal{B}$ belong to the same types, while they should form negative pairs when belonging to different types. The contrastive loss in trigger representation is:
\begin{equation} 
    \begin{aligned}
    & \mathcal{L}_{trig} = \frac{1}{n-1}\sum_{i \neq l}^{|\mathcal{B}|} - \frac{1}{m} \sum_{j \neq k}^{|\mathcal{O}|} [\mathbf{y}_i^j=\mathbf{y}_l^k] \\
    & \log  \frac{\exp(S({\mathbf{h}_e}_i^j, {\mathbf{h}_e}_l^k)/ \tau)}{\sum_{p\neq i}^{|\mathcal{B}|} \sum_{q}^{|\mathcal{O}|}[\mathbf{y}_i^j\neq \mathbf{y}_p^q]\exp(S({\mathbf{h}_e}_i^j,{\mathbf{h}_e}_p^q)/\tau)}
    \end{aligned}
\end{equation}

\subsection{Knowledge Distillation}
Similar to \citet{cao-etal-2020-incremental}, we use Knowledge Distillation at feature-level and predict-level. At task $T_t$, we distill knowledge from $T_{t-1}$ .

\textbf{Feature-level Distillation}. We get previously and currently normalized representations $\Tilde{\mathbf{h}}$ and $\mathbf{h}$ at the last layer's hidden states. We measure the similarity by function $S(\cdot)$ (\emph{Cosine Similarity}). The feature-level distillation loss is:
\begin{equation} 
%这个公式不知道写的有没有问题
    \mathcal{L}_{fd} = \sum_{(X,{Y}) \in T_t} 1 - S(\Tilde{\mathbf{h}}, \mathbf{h})
\end{equation}

\textbf{Predict-level Distillation}. As is demonstrated in \citet{hinton2015distilling}, given trigger representations $h_e$, we obtain probability distribution:

\begin{equation} 
    p(\mathbf{y}_i|\mathbf{h}_e) = \frac{\exp(\mathbf{W}_i^T \mathbf{h}_e + \mathbf{b}_i) /\tau_d}{\sum_{j \in L_{t-1}} \exp(\mathbf{W}_j^T \mathbf{h}_e + \mathbf{b}_j) /\tau_d}
\end{equation}
where $\tau_d$ is the temperature to control the smoothness of the distribution target. 
% Since the output of the teacher model on previous classification distribution has no preference for new types, thus it is not influenced by the parameter updates on new tasks. For the sake of this,
We compute previous and current probability distribution $\Tilde{\mathbf{p}}$ and $\mathbf{p}$ on previous label set $L_{t-1}$. The training objective is:
\begin{equation} 
    \mathcal{L}_{pd} = - \sum_{(X,Y) \in T_t} \Tilde{\mathbf{p}} \log \mathbf{p}
\end{equation} 
\subsection{Training}

We present detailed training procedures in Algorithm \ref{alg:train}.
In view that $\mathcal{L}_{ce}$ is the primary training objective and $\mathcal{L}_{cls}$ plays an auxiliary role to help exploit sentence information, we enable $\mathcal{L}_{ce}$ and $\mathcal{L}_{cls}$ in $T_{base}$. In $T_{inc}$, we incorporate the distillation losses ($\mathcal{L}_{fd}$ and $\mathcal{L}_{pd}$) and the exemplar replay loss ($\mathcal{L}_{re}$) as they rely on previous knowledge for training. We exclusively enable $\mathcal{L}_{trig}$ in $T_{inc}$ due to its superior effectiveness in few-shot learning. Each loss function is weighted by a factor $\lambda_i$, where $i \in \{ce, re, cls, trig, fd, pd\}$.

\section{Experiments}
\subsection{Continual Few-shot Event Detection Benchmarks}
 We construct our benchmarks based on two publicly available datasets: 
 
 \textbf{MAVEN} \citeplanguageresource{wang-etal-2020-maven}: The original MAVEN dataset contains 168 event types, which is a massive general domain event detection dataset. Regarding the training/validation/testing split, similar to \citet{yu-etal-2021-lifelong}, the test set is built upon the initial development set. We randomly select samples in the original training set to collect another development set. For incremental task split, we select the most frequent types to construct CFED tasks. Accordingly, we randomly sample 100 instances for each type in the base task, and 5 or 10 instances for each type in the incremental task.
 
 \textbf{ACE 2005} \citeplanguageresource{ace}: The ACE 2005 dataset consists of 33 event types. The training/validation/testing split is formed by previously mentioned works \citep{yang-mitchell-2016-joint,nguyen-etal-2016-joint-event}. We execute the identical operation on the incremental task split as we do on the MAVEN dataset to construct CFED tasks.
 
 Our experiments contain 5 sub-tasks. We define the task containing $m$ event types for each sub-task and $k$ training samples for each type as $m$-$way$ $k$-$shot$ CFED task. We select 10 and 20 most frequent types to conduct 2-way 5-shot, 2-way 10-shot, 4-way 5-shot and 4-way 10-shot tasks. We randomly sample 100 instances for each type in $T_{base}$, 5 and 10 instances for each type in $T_{inc}$.

\subsection{Evaluation Metrics}
Following \citet{cao-etal-2020-incremental}, we use \textbf{\textit{micro F1}} score to evaluate the performance under each stage. For stage $C_i^{test}$ we calculate ${F1}_i$ on all observed event types, as is defined in section \ref{section:definition}. \textbf{\textit{Micro F1}} score enables a comprehensive evaluation of the prediction results for all categories. We define $\bar{F1}_{micro}=\sum_{i=1}^n F1_i$ as the metric for overall performance on CFED.

\subsection{Baseline Systems}

\textbf{Fine-tune.} We fine-tune BERT continually on every sub-task. Typically, this option is the lower boundary in Continual Learning.

\textbf{Combined Retrain}. We retrain the model by combining all training samples of currently known types every time a new task arrives. It is usually regarded as the upperbound. 

\textbf{EWC} \citep{kirkpatrick2017overcoming}, which is an regularization-based method. It applies a regularization term to restrict updates for parameters that are important for previous task.

\textbf{LwF} \citep{li2017learning}, which contains a distillation module to match the probability of previous models to maintain previous knowledge.

\textbf{ICaRL} \citep{rebuffi2017icarl}, which is a memory-based method. Besides, they utilize a representation learning method.

\textbf{KCN} \citep{cao-etal-2020-incremental}, which is a popular continual event detection method following the memory replay-knowledge distillation paradigm.

\textbf{KT} \citep{yu-etal-2021-lifelong}. It generally follows the memory-based paradigm with a novel initialization method to transfer knowledge. 

\textbf{EMP} \citep{liu-etal-2022-incremental}. Besides memory replay, it introduces prompt learning of each event type to load previous types' knowledge.

\subsection{Implementation Details}
All baselines are implemented in the same settings as follows. BERT model is the open-sourced 110M bert-base-uncased from HuggingFace\footnote{\url{https://huggingface.co/bert-base-uncased}}. The number of training iterations is 30, the batch size is 4, AdamW\citep{LoshchilovH19} is used as the optimizer, the learning rate is set to 2e-5, and the weight decay is set to 1e-4. The memory capacity is 1 for each type.All computations are performed on the NVIDIA GeForce RTX 3090 (24GB) platform with 5 different random seeds. More detailed implementations can be seen in the open-sourced code repository. 

\begin{table*}[t]
    \centering
    \resizebox{.9\textwidth}{!}{
    \begin{tabular}{l|c c c c c c|c c c c c c}
    \toprule
    \multirow{2}{*}{\textbf{Method}} & \multicolumn{6}{c|}{\textbf{4-way 5-shot}} & \multicolumn{6}{c}{\textbf{4-way 10-shot}}\\
    \cmidrule{2-13}
    & 1 & 2 & 3 & 4 & 5 & $\bar{F1}_{micro}$ & 1 & 2 & 3 & 4 & 5 & $\bar{F1}_{micro}$ \\ 
    \midrule
    Fine-tune & 40.43±2.34 & 33.17±3.55 & 17.5±2.07 & 19.72±0.92 & 21.01±0.87 & 26.36±1.3 & 40.43±2.34 & 38.18±2.83 & 20.46±1.11 & 20.35±2.19 & 23.57±1.01 & 28.6±0.92 \\
    Retrain & 40.43±2.34 & 42.1±1.13 & 39.61±1.12 & 43.03±1.56 & \textbf{47.43±0.67} & 42.52±0.7 & 40.43±2.34 & 44.27±1.36 & 44.76±1.37 & \textbf{48.28±1.43} & \textbf{53.66±0.97} & 46.28±0.95 \\
    \midrule
    EWC & 40.43±2.34 & 34.29±1.41 & 17.4±1.5 & 18.61±2.52 & 20.43±1.67 & 26.23±1.39 & 40.43±2.34 & 36.42±3.34 & 19.69±0.93 & 20.02±1.14 & 23.72±1.19 & 28.06±1.01 \\
    LwF & 40.43±2.34 & 37.27±4.9 & 26.69±4.07 & 24.7±1.47 & 30.54±1.43 & 31.93±2.05 & 40.43±2.34 & 41.09±2.8 & 31.89±0.57 & 30.57±1.09 & 34.43±2.08 & 35.68±0.69 \\
    ICaRL & 35.82±4.76 & 37.16±4.85 & 33.74±2.85 & 35.54±2.37 & 35.98±2.48 & 35.65±2.93 & 35.82±4.76 & 42.43±4.48 & 37.45±1.58 & 40.11±0.9 & 41.04±1.17 & 39.37±2.05 \\
    KCN & 40.43±2.35 & 48.38±1.66 & 41.99±2.01 & 41.32±1.53 & 40.29±1.51 & 42.48±1.49 & 40.43±2.35 & 51.15±1.19 & 45.22±1.22 & 44.31±0.69 & 44.47±1.51 & 45.12±1.09 \\
    KT & 41.04±1.59 & 40.19±2.17 & 35.21±1.34 & 32.69±0.78 & 33.77±0.58 & 36.58±1.06 & 41.04±1.59 & 44.39±0.91 & 40±1.3 & 39.42±0.33 & 37.87±0.95 & 40.54±0.58 \\
    EMP & 40.17±1.34 & 30.95±0.75 & 31.21±1.32 & 22.9±2.09 & 22.25±1.43 & 29.5±0.76 & 40.17±1.34 & 32.33±0.69 & 32.95±1.11 & 26.68±1.5 & 28.16±1.89 & 32.06±0.8 \\
    \midrule
    HANet(Ours) & \textbf{41.91±3.76} & \textbf{51.39±1.55} & \textbf{43.21±3.19} & \textbf{43.53±4.21} & 43.89±5.65 & \textbf{44.79±2.33} & \textbf{41.91±3.76} & \textbf{53.17±1.27} & \textbf{46.71±2.51} & 46.36±3.64 & 48.12±5.49 & \textbf{47.25±2.23} \\
    \bottomrule
    \end{tabular}
    }
    \caption{${F1}_{micro}$ of every sub-task and $\bar{F1}_{micro}$ across all sub-tasks on 4-way MAVEN benchmark.}
    \label{tab:4wMAVEN}
\end{table*}

\begin{table*}[htb]
    \centering
    \resizebox{.9\textwidth}{!}{
    \begin{tabular}{l|c c c c c c|c c c c c c}
    \toprule
    \multirow{2}{*}{\textbf{Method}} & \multicolumn{6}{c|}{\textbf{2-way 5-shot}} & \multicolumn{6}{c}{\textbf{2-way 10-shot}}\\
    \cmidrule{2-13}
    & 1 & 2 & 3 & 4 & 5 & $\bar{F1}_{micro}$ & 1 & 2 & 3 & 4 & 5 & $\bar{F1}_{micro}$ \\ 
    \midrule
    
    Fine-tune & 60.86±2.96 & 52.09±9.59 & 46.37±10 & 26.64±6.98 & 23.15±4.66 & 41.82±3.56 & 60.86±2.96 & 48.17±9.8 & 49.55±2.91 & 23.29±8.2 & 24.66±3.23 & 41.31±3.31 \\
    Retrain & 60.86±2.96 & 62.45±4.27 & 52.21±7.83 & 52.2±4.68 & \textbf{58.36±6.09} & 57.22±4.48 & 60.86±2.96 & 63.39±2.87 & 63.75±2.67 & \textbf{61.23±2.08} & \textbf{64.25±3.13} & \textbf{62.7±1.3} \\
    \midrule
    EWC & 60.86±2.96 & 49.3±8.93 & 45.41±10.43 & 27.14±11.24 & 22.36±3.9 & 41.02±4.85 & 60.86±2.96 & 47.58±10.11 & 51.15±3.05 & 23.82±7.67 & 21.79±3.1 & 41.04±2.78 \\
    LwF & 60.86±2.96 & 47.31±10.4 & 38.91±12.89 & 23.31±13.46 & 28.4±2.83 & 39.76±6.85 & 60.86±2.96 & 46.98±8.32 & 50.77±3.35 & 33.48±2.7 & 29.69±2.91 & 44.36±2.2 \\
    ICaRL & 50.85±6.51 & 52.21±2.72 & 37.39±6.78 & 31.33±6.31 & 28.85±5.04 & 40.13±4.1 & 50.85±6.51 & 52.06±2.66 & 42.45±6.48 & 32.89±4.96 & 34.7±3.93 & 42.59±2.8 \\
    KCN & 60.86±2.96 & 56.38±5.03 & 47.56±10.41 & 38.62±9.47 & 37.05±7.11 & 48.09±6.41 & 60.86±2.96 & 59.41±6.74 & 57.39±6.19 & 46.48±6.1 & 44.3±5.43 & 53.69±4.42 \\
    KT & 53.16±2.25 & 42.55±2.33 & 33.93±2.97 & 38.48±8.66 & 31.27±9.34 & 39.88±3.84 & 53.16±2.25 & 59.12±1.78 & 50.02±5.13 & 49.02±5.34 & 28.54±2.95 & 47.97±2.67 \\
    EMP & 54.78±1.49 & 40.49±1.9 & 24.32±3.37 & 27.15±8.46 & 22.53±6.02 & 33.85±2.96 & 54.78±1.49 & 37.28±7.37 & 19.6±4.96 & 34.69±4.76 & 24.19±6.62 & 34.11±3.48 \\
    \midrule
    HANet(Ours) & \textbf{61.16±2.29} & \textbf{63.07±3.09} & \textbf{57.5±5.98} & \textbf{53.21±4.64} & 54.31±3.21 & \textbf{57.85±2.91} & \textbf{61.16±2.29} & \textbf{66.84±2.88} & \textbf{64.68±3.77} & 58.02±6.58 & 54.37±5.94 & 61.02±3.46 \\
    \bottomrule
    \end{tabular}
    }
    \caption{${F1}_{micro}$ of every sub-task and $\bar{F1}_{micro}$ across all sub-tasks on 2-way ACE benchmark.}
    \label{tab:2wACE}
\end{table*}

\subsection{Main Results}
\begin{figure}[t]
    \centering
    \includegraphics[width=1.0\linewidth]{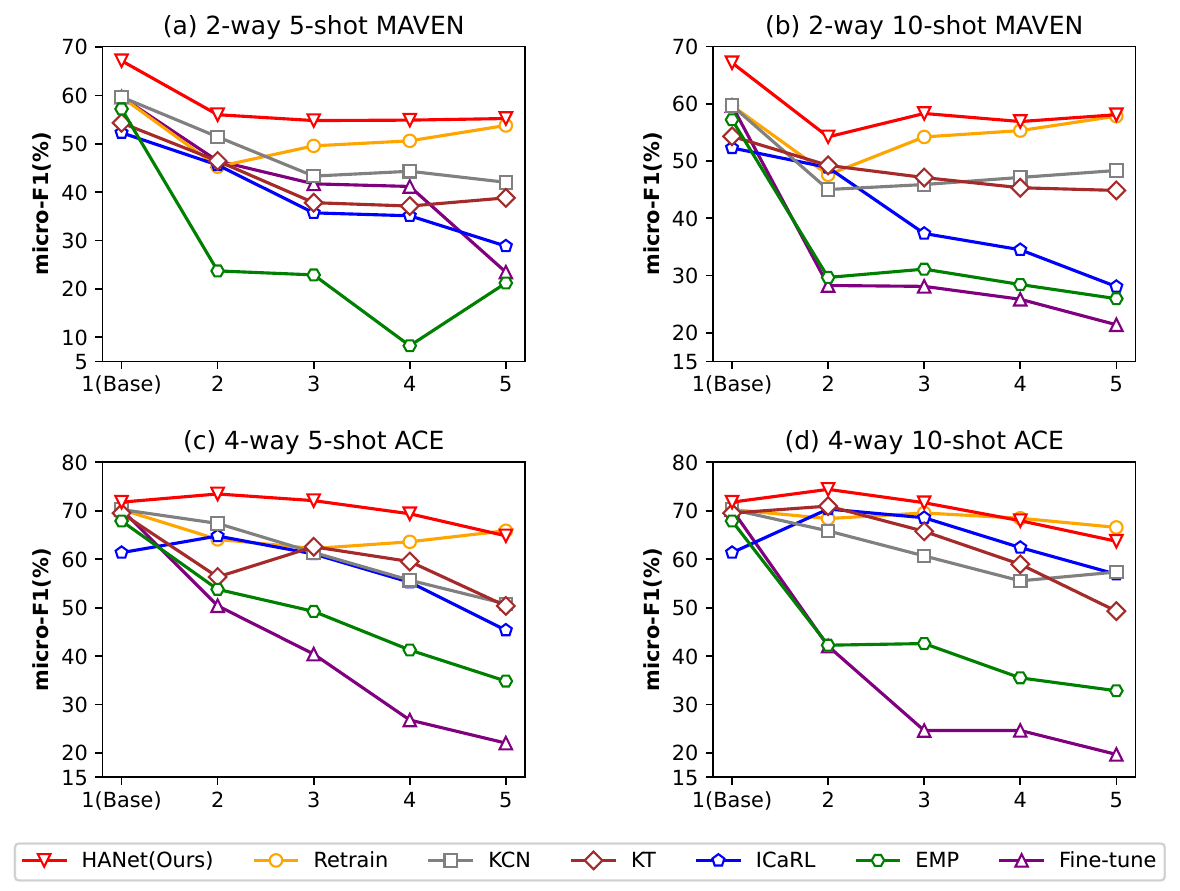}
    \caption{${F1}_{micro}$ performance of every sub-task on 2-way MAVEN and 4-way ACE.}
    \label{fig:4wACE}
\end{figure}

We conduct each experiment 5 times and report the $\mathbf{means\pm std.}$ on MAVEN and ACE benchmarks in comparison with previously mentioned baselines. We report results in Table \ref{tab:4wMAVEN}, and Table \ref{tab:2wACE} and Figure \ref{fig:4wACE}. From the results, we can observe that:

(1) Compared with previous baselines, our approach significantly outperforms them across all sub-tasks. On 4-way 5-shot MAVEN and 2-way 5-shot ACE, our model obtains improvements of $7.27\%$ and $8.44\%$ on $\bar{F1}_{micro}$ when compared with previous state-of-the-art methods. Our approach even exceeds the strong retrain baseline with improvements of $5.94\%$ and $5.56\%$ on $\bar{F1}_{micro}$, which strongly proves the effectiveness of our approach.

(2) KCN and KT achieve relatively good performance. As we limit the memory capacity to only one sample for each type to replay, they can learn little knowledge from memory replay, which strongly demonstrates the importance of characterizing prototypical feature space. 

\begin{table*}[ht]
  \centering
    \resizebox{.9\textwidth}{!}{
    \begin{tabular}{l|cccccc|cccccc}
    \toprule
    \multicolumn{1}{c|}{\multirow{2}[3]{*}{\textbf{Method}}} & \multicolumn{6}{c|}{\textbf{2-way 5-shot}} & \multicolumn{6}{c}{\textbf{2-way 10-shot}} \\
\cmidrule{2-13}      & 1 & 2 & 3 & 4 & 5 & $\bar{F1}_{micro}$ & 1 & 2 & 3 & 4 & 5 & $\bar{F1}_{micro}$ \\
    \midrule
    HANet(Ours) & \textbf{67.16 } & \textbf{56.01 } & \textbf{54.80 } & \textbf{54.89 } & \textbf{55.22 } & \textbf{57.62 } & \textbf{67.16 } & \textbf{54.22 } & \textbf{58.31 } & \textbf{56.90 } & \textbf{58.09 } & \textbf{58.94 } \\
    w/o Replay* & \textbf{67.16 }  & 51.02  & 44.15  & 38.76  & 36.78  & 47.57  & \textbf{67.16 }  & 48.13  & 48.14  & 41.07  & 40.01  & 48.90  \\
    w/o Distill & \textbf{67.16 }  & 46.83  & 42.77  & 37.17  & 42.90  & 47.37  & \textbf{67.16 }  & 45.45  & 44.07  & 44.90  & 47.77  & 49.87  \\
    w/o PA & \textbf{67.16 }  & 54.28  & 53.01  & 50.98  & 52.21  & 55.53  & \textbf{67.16 }  & 52.94  & 57.47  & 53.91  & 55.38  & 57.37  \\
    w/o CA & 59.67  & 54.45  & 49.14  & 50.08  & 49.57  & 52.58  & 59.67  & 53.31  & 53.75  & 53.16  & 53.46  & 54.67  \\
    w/o PA and CA & 59.67  & 51.43  & 43.32  & 44.32  & 42.04 & 48.16  & 59.67  & 45.03  & 45.90  & 47.14  & 48.35  & 49.22 \\

    \bottomrule
    \end{tabular}
    }
  \caption{We perform ablation studies, comparing ${F1}_{micro}$ by removing each component at a time.}
  \label{tab:ablation}
\end{table*}

\begin{table*}[htbp]
    \centering
    \resizebox{.9\textwidth}{!}{
        \begin{tabular}{c|l|cccccc|cccccc}
        \toprule
        \multirow{2}[2]{*}{\textbf{Benchmark}} & \multicolumn{1}{c|}{\multirow{2}[2]{*}{\textbf{Method}}} & \multicolumn{6}{c|}{\textbf{2-way 1-shot}} & \multicolumn{6}{c}{\textbf{2-way 2-shot}} \\
        \cmidrule{3-14}
          &   & 1 & 2 & 3 & 4 & 5 & $\bar{F1}_{micro}$ & 1 & 2 & 3 & 4 & 5 & $\bar{F1}_{micro}$ \\
        \midrule
        \multirow{2}[2]{*}{MAVEN} & HANet(Ours) & \textbf{67.16 } & 45.54  & 38.28  & \textbf{42.39 } & \textbf{40.40 } & \textbf{46.75 } & \textbf{67.16 } & 55.87  & \textbf{50.35 } & \textbf{51.63 } & \textbf{51.39 } & \textbf{55.28 } \\
          & gpt-3.5-turbo & 54.22  & \textbf{55.25 } & \textbf{41.60 } & 37.88  & 33.31  & 44.45  & 57.00  & \textbf{58.51 } & 43.64  & 40.39  & 36.56  & 47.22  \\
        \midrule
        \multirow{2}[2]{*}{ACE} & HANet(Ours) & \textbf{60.99 } & \textbf{51.93 } & \textbf{41.67 } & 41.54  & \textbf{35.84 } & \textbf{46.40 } & \textbf{60.99 } & \textbf{58.38 } & 39.48  & 41.76  & \textbf{44.60 } & \textbf{49.04 } \\
          & gpt-3.5-turbo & 42.20  & 50.29  & 40.51  & \textbf{43.46 } & 35.21  & 42.33  & 56.36  & 49.72  & \textbf{45.16 } & \textbf{44.44 } & 42.96  & 47.73  \\
        \bottomrule
        \end{tabular}%
    }
    \caption{Comparison with gpt-3.5-turbo on MAVEN and ACE benchmark.}
    \label{tab:llm}
\end{table*}%

\begin{table}[t]
    \centering
    \resizebox{.9\linewidth}{!}{
    \begin{tabular}{c|l|cc|cc}
    \toprule
     \multirow{2}{*}{\textbf{Way-num}}  & \multirow{2}{*}{\textbf{Method}} & \multicolumn{2}{c|}{\textbf{MAVEN}} & \multicolumn{2}{c}{\textbf{ACE}} \\
    &   & 5-shot & \multicolumn{1}{c|}{10-shot} & 5-shot & 10-shot \\
    \midrule
    \multirow{5}[4]{*}{2way} 
    & w/o CA & 52.58 & \multicolumn{1}{c|}{54.67} & 48.27 & 60.45 \\
    & Dropout & 54.68 & \multicolumn{1}{c|}{56.32} & 53.06 & 61.87 \\
    & Shuffle & \textbf{57.62} & \multicolumn{1}{c|}{\textbf{58.94}} & 55.10 &   \textbf{63.98} \\
    & RTR & 54.60 & \multicolumn{1}{c|}{56.57} & \textbf{55.53} & 63.27 \\
    & Retrain & 51.78  & \multicolumn{1}{c|}{54.93 } & 49.54  & 60.69  \\
    \midrule   
    \multirow{5}[3]{*}{4way} 
    & w/o CA & 45.68 & \multicolumn{1}{c|}{48.95} & 64.66 & 68.70 \\
    & Dropout & 44.36 & \multicolumn{1}{c|}{47.45} & 67.41 & 68.58 \\
    & Shuffle & \textbf{48.47} & \multicolumn{1}{c|}{\textbf{49.91}} & \textbf{70.31} & \textbf{69.90} \\
    & RTR & 46.18 & \multicolumn{1}{c|}{47.96} & 67.93 & 68.11 \\
    & Retrain & 42.53  & 46.59  & 65.21  & 68.65  \\
     \bottomrule
    \end{tabular}
    }
  \caption{${\bar{F1}_{micro}}$ of different augmentation methods on MAVEN and ACE benchmarks. We also list the ``w/o CA'' and Retrain method for comparison.}
  \label{tab:sample}

\end{table}

(3) When compared with methods optimized for continual event detection, traditional methods: EWC, LwF, and ICaRL perform poorly. The giant gap between the lower bound and HANet illustrates that CFED is a challenging task.

\subsection{Ablation Study}
\begin{figure}[t]
    \centering
    \includegraphics[width=1.0\linewidth]{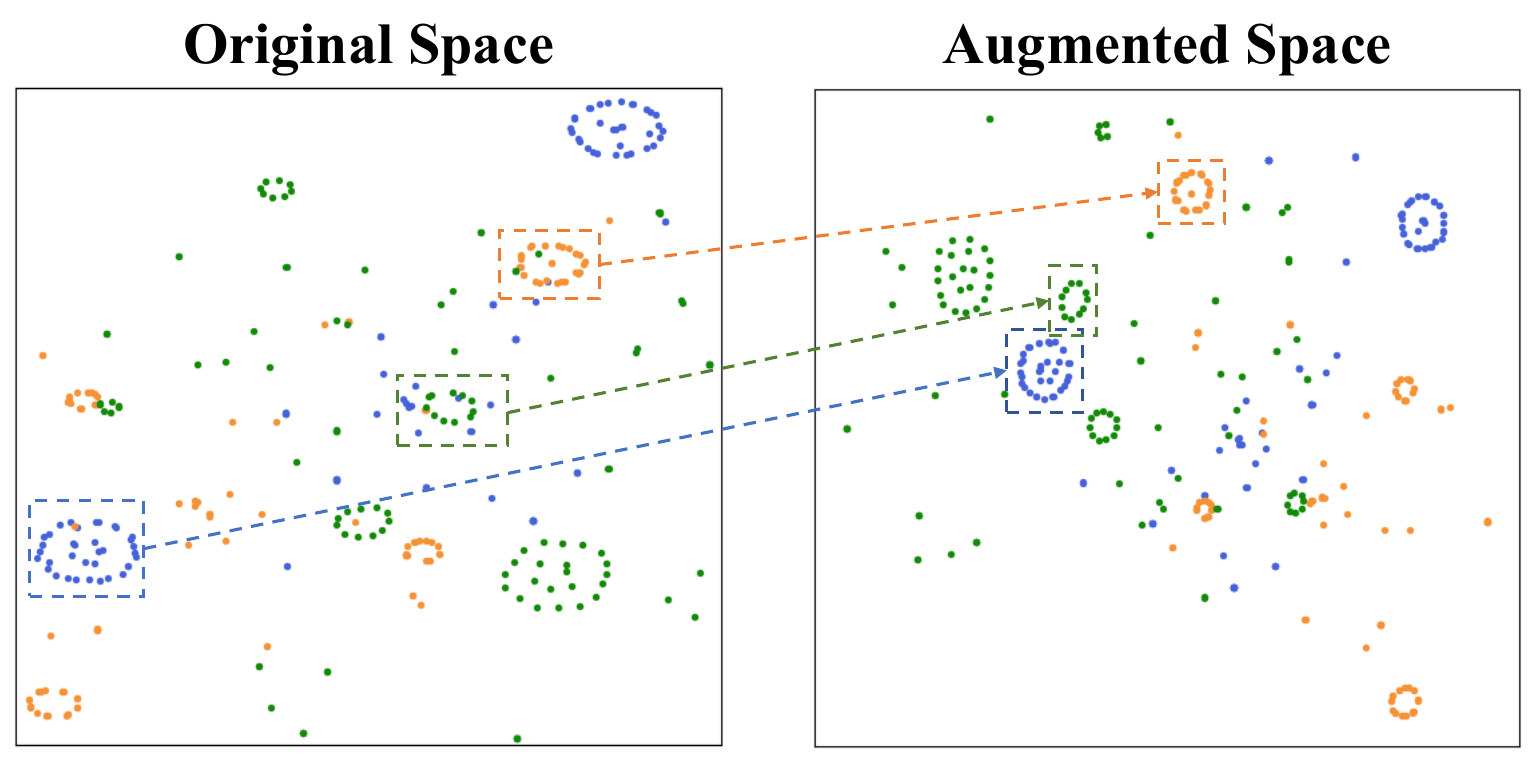}
    \caption{Embedding space visualization via t-SNE on original and prototypical augmented feature in task $T_2$. Points within the same color indicate identical event types. As we can see, after prototypical augmentation, the intra-class distances become closer for each type. Besides, some hard samples (pointed in the squared region) initially proximate to the centers of other classes in the original space become easier to classify after prototypical augmentation, showcasing the effectiveness of prototypical augmentation.}
    \label{visulize}
\end{figure}

We conduct ablation study to validate the effectiveness of each component. We choose 2-way MAVEN for the ablation study in Table \ref{tab:ablation}. The ``Replay*'' denotes removing memory replay. As prototypical augmentation is based on memory set, $\mathcal{L}_{re}$ is also set to $0$ in ``Replay*''. The distillation losses $\mathcal{L}_{fd}$ and $\mathcal{L}_{pd}$ are removed in ``w/o Distill''. $\mathcal{L}_{re}$ and $\mathcal{L}_{cls}$ and $\mathcal{L}_{trig}$ are removed in settings ``w/o PA'' and ``w/o CA'', respectively. Here are the conclusions:

(1) \textbf{Effectiveness of Prototypical Augmentation.} Compared with removing prototypical augmentation (PA), PA boosts the performance by an average of $2.09\%$ and $1.57\%$. Meanwhile, with the task proceeding, the model can gain more improvements, demonstrating that PA plays an increasingly vital effect in alleviating catastrophic forgetting. We also plot t-SNE visualization in Figure \ref{visulize} to show how PA contributes to memorizing previous event types.  

(2) \textbf{Effectiveness of Contrastive Augmentation.} In comparison with removing contrastive augmentation, our approach delivers improvements of $5.04\%$ and $4.27\%$ on $\bar{F1}_{micro}$, which indicates that contrastive augmentation is beneficial in mitigating overfitting in few-shot incremental tasks. Although we focus more on on $T_{inc}$, the model can greatly benefit from the auxiliary objectives in $T_{base}$. 

(3) \textbf{Effectiveness of  Prototypical Augmentation and Contrastive Augmentation.} When removing prototypical augmentation and contrastive augmentation, the $\bar{F1}_{micro}$ faces a sharp decline of $9.46\%$ and $9.72\%$, implying the synergistic effect of the two modules to address the CFED problem.

\subsection{Effect of Augmentation Method in Contrastive Augmentation}

Different augmentation methods affect contrastive augmentation. We evaluate ``Dropout'', ``Shuffle'', and ``Random Token Replacement'' (``RTR''). As mentioned in \citet{gao-etal-2021-simcse}, ``Dropout'' means making a forward pass with dropout modules. ``Shuffle'' randomly shuffle the sentence. ``RTR'' refers to randomly replacing non-trigger tokens with other tokens. From Table \ref{tab:sample}, we can draw the following conclusion: In most cases, ``Shuffle'' is the most effective method. ``Dropout'' performs worse than the others, however, it still outperforms ``w/o CA''.

\begin{figure}[tbp]
    \centering
    \includegraphics[width=1.0\linewidth]{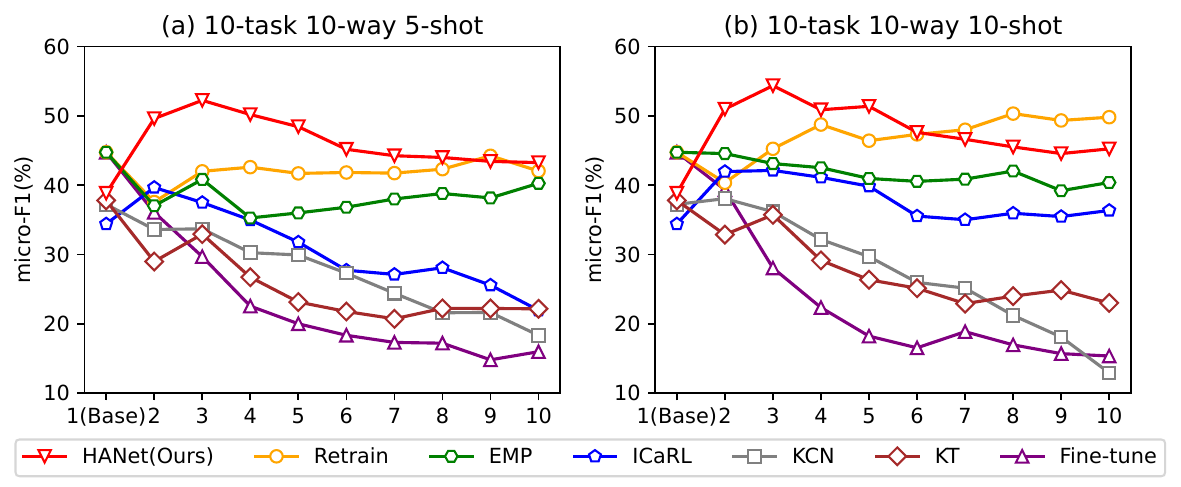}
    \caption{${F1}_{micro}$ performance of each sub-task in Larger MAVEN benchmark.}
    \label{fig:larger}
  
\end{figure}

\begin{table*}[t]
    \centering

    \resizebox{.9\textwidth}{!}{
    \begin{tabular}{l|c c c c c c|c c c c c c}
    \toprule
    \multirow{2}{*}{\textbf{Method}} & \multicolumn{6}{c|}{\textbf{2-way 1-shot}} & \multicolumn{6}{c}{\textbf{2-way 2-shot}}\\
    \cmidrule{2-13}
    & 1 & 2 & 3 & 4 & 5 & $\bar{F1}_{micro}$ & 1 & 2 & 3 & 4 & 5 & $\bar{F1}_{micro}$ \\ 
    \midrule
    Fine-tune & 59.67  & 26.81  & 28.34  & 22.96  & 18.79  & 31.31  & 59.67  & 56.17  & 41.67  & 33.13  & 22.81  & 42.69  \\
    Retrain & 59.67  & 42.34  & 33.33  & 29.04  & 28.25  & 38.53  & 59.67  & 44.68  & 37.73  & 38.70  & 40.98  & 44.35  \\
    \midrule
    EWC & 59.67  & 35.95  & 28.22  & 15.79  & 16.17  & 31.16  & 59.67  & 55.68  & 47.96  & 36.10  & 26.92  & 45.27  \\
    LwF & 59.67  & 5.28  & 24.63  & 27.11  & 30.82  & 29.50  & 59.67  & 36.72  & 34.07  & 28.94  & 28.71  & 37.62  \\
    ICaRL & 52.29  & 36.71  & 34.18  & 31.06  & 25.77  & 36.00  & 52.29  & 41.38  & 34.44  & 33.47  & 29.19  & 38.15  \\
    KCN & 59.67  & 39.10  & \textbf{43.19 } & 41.97  & 38.18  & 44.42  & 59.67  & 54.40  & \textbf{50.67 } & 49.98  & 47.58  & 52.46  \\
    KT & 54.32  & 5.94  & 5.78  & 3.70  & 3.61  & 14.67  & 54.32  & 35.22  & 32.71  & 27.47  & 28.23  & 35.59  \\
    EMP & 57.21  & 4.95  & 5.53  & 5.42  & 5.29  & 15.68  & 57.21  & 18.28  & 6.84  & 7.06  & 8.43  & 19.56  \\
    \midrule
    HANet(Ours) & \textbf{67.16 } & \textbf{45.54 } & 38.28  & \textbf{42.39 } & \textbf{40.40 } & \textbf{46.75 } & \textbf{67.16 } & \textbf{55.87 } & 50.35  & \textbf{51.63 } & \textbf{51.39 } & \textbf{55.28 } \\
    \bottomrule
    \end{tabular}
    }
    \label{tab:fewer2wMAVEN}
    \caption{2-way Continual Fewer-shot Event Detection Task in MAVEN benchmark.}
    \label{tab:fewerMAVEN}
\end{table*}

\begin{table*}[ht]
    % \begin{subtable}{1.0\linewidth}
        \centering
    \resizebox{.9\textwidth}{!}{
    \begin{tabular}{l|c c c c c c|c c c c c c}
    \toprule
    \multirow{2}{*}{\textbf{Method}} & \multicolumn{6}{c|}{\textbf{2-way 1-shot}} & \multicolumn{6}{c}{\textbf{2-way 2-shot}}\\
    \cmidrule{2-13}
    & 1 & 2 & 3 & 4 & 5 & $\bar{F1}_{micro}$ & 1 & 2 & 3 & 4 & 5 & $\bar{F1}_{micro}$ \\ 
    \midrule
    Fine-tune & 57.75  & 52.97  & 26.47  & 15.87  & 3.50  & 31.31  & 57.75  & 49.38  & 26.01  & 22.52  & 29.71  & 37.07  \\
    Retrain & 57.75  & 43.16  & 30.16  & 31.69  & 28.36  & 38.22  & 57.75  & 48.91  & 33.86  & 36.97  & 35.01  & 42.50  \\
    \midrule
    EWC & 57.75  & 45.09  & 25.37  & 16.18  & 6.51  & 30.18  & 57.75  & 50.60  & 23.87  & 13.90  & 25.46  & 34.32  \\
    LwF & 57.75  & 37.50  & 18.31  & 7.97  & 6.37  & 25.58  & 57.75  & 44.00  & 16.72  & 16.29  & 29.45  & 32.84  \\
    ICaRL & 54.68  & 45.96  & 27.08  & 25.29  & 22.34  & 35.07  & 54.68  & 43.81  & 32.89  & 33.12  & 28.49  & 38.60  \\
    KCN & 57.75  & 54.13  & 40.71  & \textbf{43.97 } & 26.52  & 44.61  & 57.75  & 51.37  & 36.83  & 34.66  & 40.40  & 44.20  \\
    KT & 51.90  & 1.47  & 1.36  & 1.14  & 1.51  & 11.48  & 51.90  & 40.19  & 24.03  & 24.20  & 20.81  & 32.22  \\
    EMP & 56.10  & 1.77  & 3.59  & 3.59  & 3.70  & 13.75  & 56.10  & 34.11  & 16.57  & 3.62  & 15.09  & 25.10  \\
    \midrule
    HANet(Ours) & \textbf{60.99 } & \textbf{51.93 } & \textbf{41.67 } & 41.54  & \textbf{35.84 } & \textbf{46.40 } & \textbf{60.99 } & \textbf{58.38 } & \textbf{39.48 } & \textbf{41.76 } & \textbf{44.60 } & \textbf{49.04 } \\
    \bottomrule
    \end{tabular}
    }
    \caption{2-way Continual Fewer-shot Event Detection Task in ACE benchmark.}
    \label{tab:fewer2wACE}

\end{table*}

\subsection{Evaluation in Extreme Scenarios}

To validate the effeciveness of our method in various CFED applications, we conduct experiments to investigate on extreme conditions with more incremental tasks and fewer shot numbers.
.
\textbf{Larger CFED Task.} We exploit MAVEN benchmark to select 100 most frequent types to conduct 10-task 10-way task. From the results in Figure \ref{fig:larger}, we conclude that existing methods can not generalize well to larger CFED, meanwhile, HANet still maintains the best performance, showcasing strong continual learning ability in more practical situations.

\textbf{Continual Fewer-shot Event Detection Task.} To explore the minimum samples from which models can learn to maintain good performance, we perform 2-way 1-shot and 2-way 2-shot experimental settings. According to Table \ref{tab:fewerMAVEN} and \ref{tab:fewer2wACE}, our method outperforms other baselines, proving the ability to better utilize few-shot samples in severe conditions when dealing with CFED tasks.

\subsection{Capability of LLM in Solving Continual Few-shot Event Detection} 

Recently, there have been growing discussions \citep{chen2023robust,wang2023instructuie} about the capabilities of Large Language Models (LLMs) on IE tasks. Though these LLMs demonstrate promising abilities to learn from few-shot samples, their performance on continual few-shot event detection is to be discussed. In this section, we aim to evaluate the capability of ChatGPT in CFED settings. We conduct comparisons with gpt-3.5-turbo\footnote{\url{https://api.openai.com/v1/chat/completions}}.

 Following Event Extraction Trigger instructions by \citet{wang2023instructuie} to perform in-context learning in gpt-3.5-turbo \citep{ouyang2022training}, we use few-shot samples as instructions selected from the training set. The original training set in $T_1$ contains 100 samples, we randomly select 1 or 2 samples every time a new test sample arrives. Specifically, at stage $C_t$, we conduct evaluations in $C_t^{test}$ by providing few-shot samples of each type in $C_t^{train}$. Detailed instructions and cases of gpt-3.5-turbo are shown in Appendix \hyperref[app:llm]{A}.

 From the results illustrated in Table \ref{tab:llm}, we can observe that, compared with gpt-3.5-turbo failed to perform well on continual few-shot event detection tasks. Our method outperforms gpt-3.5-turbo significantly.

\section{Related Work}

\subsection{Traditional Event Detection}
Impressive progress has been made in research related to traditional event detection by neural network-based methods \citep{chen-etal-2015-event,nguyen-grishman-2015-event,liu-etal-2017-exploiting,chen-etal-2018-collective,lu-etal-2019-distilling}. These approaches greatly improved the performance on the ideal ED task. Nevertheless, they face considerable catastrophic forgetting and few-shot overfitting when handling continual event types with few samples, which seriously restricts their real-world applications.
\subsection{Continual Event Detection}
The major challenge of Continual ED is to learn emerging tasks while avoiding forgetting previous tasks \citep{mccloskey1989catastrophic,ring1994continual,thrun1995lifelong,thrun1998lifelong}. \citet{cao-etal-2020-incremental} construct a replay-distillation method to preserve knowledge from memory set and previous models. Besides replay and distillation, \citet{yu-etal-2021-lifelong} utilize an initialization method to transfer knowledge. \citet{liu-etal-2022-incremental} adopt prompt learning for preserving previous knowledge. Although these works perform well on Continual ED, their abilities are limited with few-shot samples.
\subsection{Few-shot Event Detection}
Few-shot event detection aims to learn great representations with insufficient samples. \citet{lai-etal-2020-extensively} propose two matching losses to provide cluster signals for few-shot learning. \citet{deng2020meta} introduce a prototypical network with dynamic memory. \citet{zhang-etal-2022-hcl} design a hybrid contrastive learning approach. \citet{zhao-etal-2022-knowledge} align event types to FrameNet to obtain more instances for prototype calculation. Since these methods only concentrate on few-shot tasks with fixed types, they dismiss the continual situation.

\section{Conclusions}
In this paper, we focus on a more realistic yet challenging scenario of continual few-shot event detection, where the system is required to detect and classify events on continually emerging new types with limited labeled data. We propose a \textbf{H}ierarchical \textbf{A}ugmentation \textbf{Net}work (\textbf{HANet}). To alleviate catastrophic forgetting in memorizing previous event types, we incorporate prototypical augmentation to preserve previous knowledge with limited exemplars. We also devise a contrastive augmentation module to tackle with overfitting when learning new event types. This module leverages valuable token information from limited samples in incremental tasks. We conduct a series of experiments to show that our model perform well on continual few-shot event detection tasks, achieving state-of-the-art performance compared with previous baselines and ChatGPT.

\section{Limitations}
Though performing well on the CFED task, there are still some limitations to be mentioned: (1) Our method focuses on a fixed emerging number of event types and the shot number of each few-shot task is unchanging, which is still ideal in real-world scenarios. (2) Though we propose space augmentation for prototypes in memory, the approach still requires extra storage space, which limits its application in some extreme scenarios. (3) Since our method performs well for event detection, it has the potential to explore the possibility of extending our approach to other IE applications (e.g., Relation Extraction and Named Entity Recognition). We leave this as future work.

\section{Acknowledgements}
This work is supported by the National Key Research and Development Program of China (No. 2022ZD0160503), and the National Natural Science Foundation of China (No. 62176257). This work is also supported by the Youth Innovation Promotion Association CAS, and Yunnan Provincial Major Science and Technology Special Plan Projects (No.202202AD080004).

\nocite{*}
\section{Bibliographical References}\label{sec:reference}

\bibliographystyle{lrec-coling2024-natbib}
\bibliography{bibiograohical_reference}

\section{Language Resource References}
% \label{lr:ref}
\bibliographystylelanguageresource{lrec-coling2024-natbib}
\bibliographylanguageresource{languageresource}

\section*{Appendix A. Instructions for large language models \label{app:llm}}
\begin{figure}[ht]
    \centering
    \includegraphics[width=1.0\linewidth]{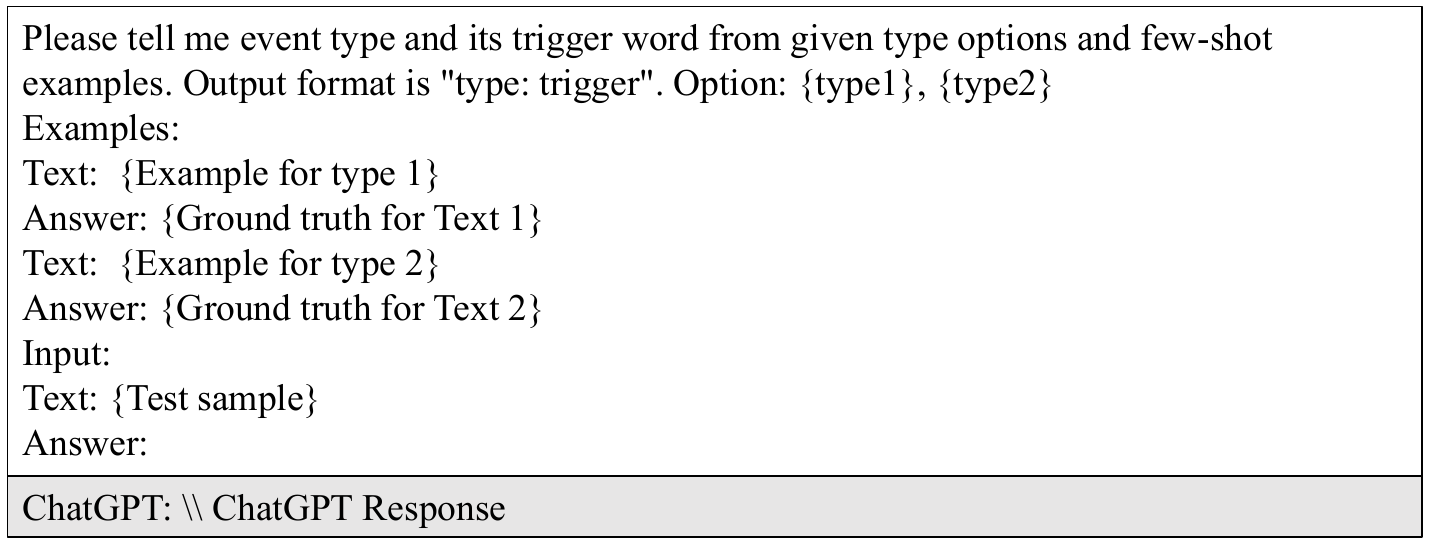}
    \caption{Instructions of gpt-3.5-turbo for CFED}
    \label{fig:instruct}
\end{figure}

\begin{figure}[ht]
    \centering
    \includegraphics[width=1.0\linewidth]{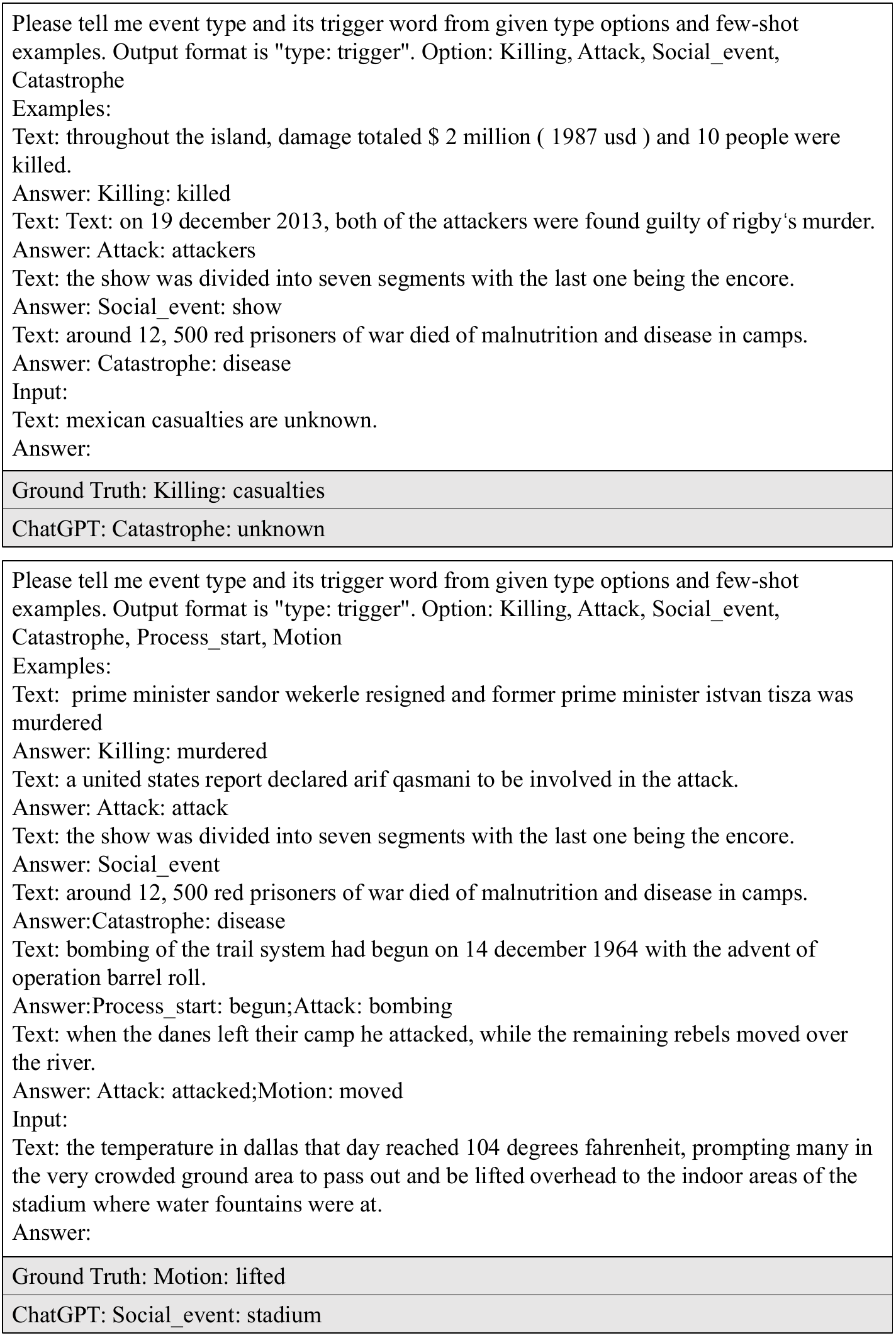}
    \caption{Cases of gpt-3.5-turbo for CFED}
    \label{fig:case}
\end{figure}

In this section, we show gpt-3.5-turbo's instructions and cases for the continual few-shot event detection task in Figure \ref{fig:instruct} and Figure \ref{fig:case}. When learning new event types, we simply append new options and examples for these types as in-context learning prompts. 

\end{document}